\documentclass[letterpaper, 10 pt, conference]{ieeeconf}
\usepackage{graphicx} 

\usepackage{subcaption}
\usepackage{colonequals}
\usepackage{amsmath} 
\usepackage{float}
\usepackage{booktabs}
\usepackage{amsfonts}
\usepackage{amssymb}
\usepackage{gensymb}
\usepackage{xcolor}
\usepackage{siunitx}

\usepackage[font=small,labelfont=bf]{caption}

\usepackage[noadjust]{cite}

\usepackage[ruled]{algorithm2e}
\makeatletter
\newcommand{\algorithmstyle}[1]{\renewcommand{\algocf@style}{#1}}
\newcommand{\removelatexerror}{\let\@latex@error\@gobble}
\makeatother

\IEEEoverridecommandlockouts                              

\title{\LARGE \bf
	Mapless Navigation among Dynamics with Social-safety-awareness: \\ a reinforcement learning approach from 2D laser scans
}

\author{Jun Jin$^{*,\dagger}$,  Nhat M. Nguyen$^{\dagger}$,  Nazmus Sakib$^{*,\dagger}$, Daniel Graves$^{\dagger}$, Hengshuai Yao$^{\dagger}$ and Martin Jagersand$^{*}$
\thanks{$^{\dagger}$Noah's Ark Lab, Huawei Technologies Canada, Ltd., Edmonton AB., Canada,	{\tt\small \{jun.jin1, minh.nhat.nguyen, daniel.graves, hengshuai.yao\}@huawei.com}}%
	\thanks{$^{*}$Department of Computing Science,
		University of Alberta, Edmonton AB., Canada
		{\tt\small \{jjin5, nazmus, mj7\}@ualberta.ca}}%
}

\begin{document}
	\maketitle
	\thispagestyle{empty}
	\pagestyle{empty}
	\begin{abstract}
	We consider the problem of mapless collision-avoidance navigation where humans are present using 2D laser scans. Our proposed method uses ego-safety to measure collision from the robot's perspective and social-safety to measure the impact of robot's actions on surrounding pedestrians. Specifically, the social-safety part predicts the intrusion impact of the robot's action into the interaction area with surrounding humans. We train the policy using reinforcement learning on a simple simulator and directly evaluate the learned policy in Gazebo and real robot tests. Experiments show the learned policy smoothly transferred to different scenarios without any fine tuning. We observe that our method demonstrates time-efficient path planning behavior with high success rate in the mapless navigation task. Furthermore, we test our method in a navigation task among dynamic crowds, considering both low and high volume traffic. Our learned policy demonstrates cooperative behavior that actively drives our robot into traffic flows while showing respect to nearby pedestrians. Evaluation videos are at https://sites.google.com/view/ssw-batman
	\end{abstract}
	\section{Introduction}\label{sec:intro}
	The problem of autonomously navigating a robot in a map-unknown (mapless) environment while avoiding both static and dynamic obstacles, is important but challenging in applications like delivery robots, indoor service robots, etc. The \textit{path planning and static obstacle avoidance} parts in this problem are often formatted as \textit{mapless navigation}~\cite{giovannangeli2006robust,tai2017virtual} where a robot is driven by observed sensory data from the unknown environment, assuming the relative pose from robot to target is given by a third party localization module (e.g., GPS for outdoor cases and UWB / Wifi / Zigbee for indoors.). The \textit{dynamic obstacle avoidance} part in this problem is more complicated since it requires future prediction on unknown surrounding dynamics like moving pedestrians, vehicles or other robots~\cite{chen2017CADRL,chen2019crowdEPFL}. As complexity of surrounding dynamics increases, the prediction may result in occupying a large portion of free space, successively causing no solution in path planning, namely the freezing robot problem~\cite{fan2019getting}. Moreover when the moving obstacles are human, not only collision avoidance, but also \textit{human-awareness}~\cite{sisbot2007human,kruse2013human} should be considered. While recent approaches using multi-modal sensing~\cite{liu2017learning} or high-end object/pedestrian detection pipelines~\cite{chen2017SACADRL,ETH2016MaxEnt,chen2019crowdEPFL} have been proposed to tackle part of the problem, we are still curious to ask: is it possible to solve for all parts purely based on 2D laser scans?
	
		\begin{figure}
		\begin{center}
			\includegraphics[width=0.4\textwidth]{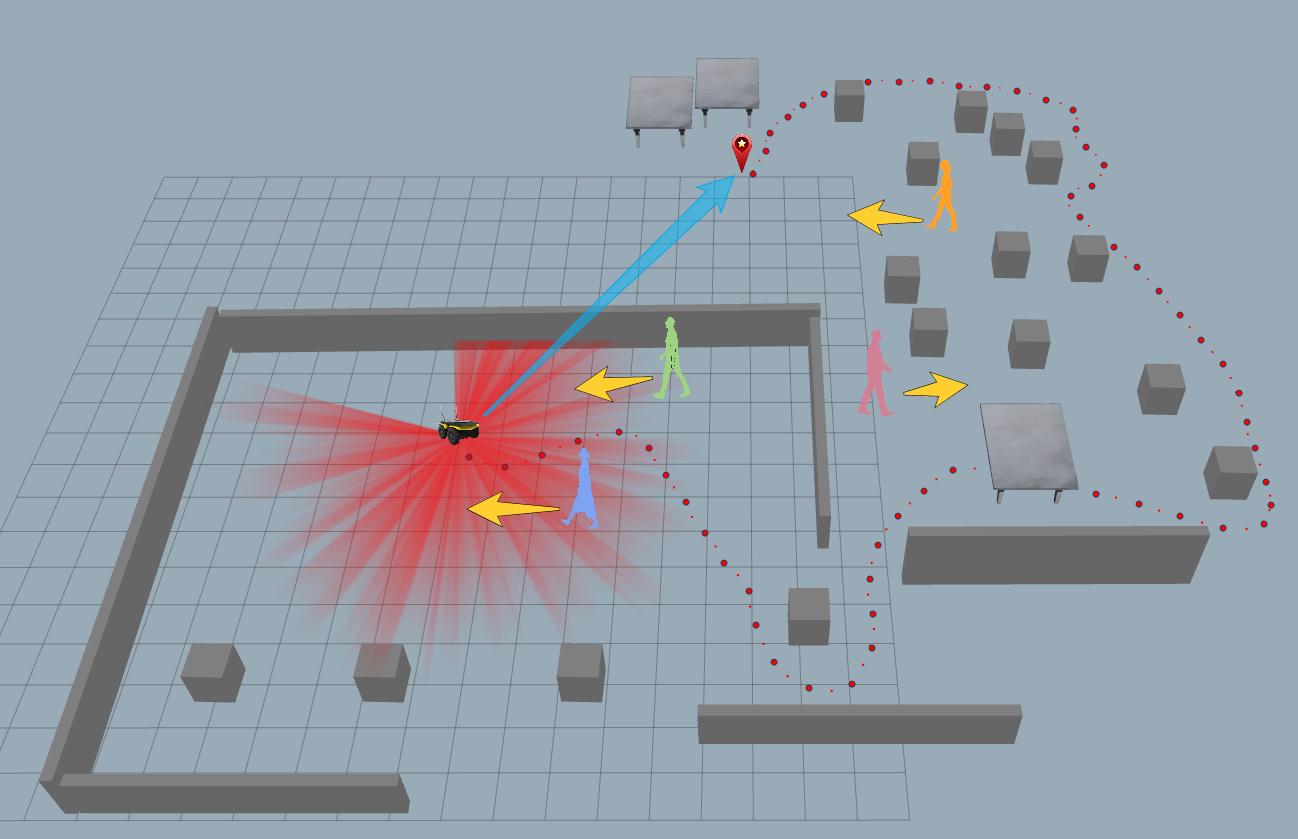} 
			\caption{Given a target location, humans can navigate locally without knowing a detailed map of surroundings. We typically make decisions from sensory inputs to avoid obstacles and walk-in pedestrians while showing respect to other people. Our decisions come from both our egocentric and allocentric spatial cognition. Can we train a robot with a similar behavior?}
			\label{fig:design_overview}
		\end{center}
	\end{figure}
	
	Recent works on deep reinforcement learning have proven the capability of using 2D laser scans in mapless navigation\cite{tai2017virtual,zhelo2018curiosity} and multi-agent / dense crowd\cite{long2018towards,fan2018crowdmove} collision avoidance tasks. Such methods share a similar reward structure by adding a reaching target reward and a collision penalty, but differ in the training process. For example, in a mapless navigation task, policy training is done by using a fixed number of floor plans\cite{tai2017virtual, zhelo2018curiosity}; In multi-agent / dense crowd\cite{long2018towards,fan2018crowdmove} collision avoidance task, the training involves interactions with randomized autonomous agents. It is observed that the collision penalty combined with sufficient exploration in training, makes the learned policy show cooperative behaviors that solve the freezing robot problem. However, this collision penalty only measures near impact of the robot from surrounding agents and obstacles. When humans are present nearby, the inverse directional impact on human safety should also be considered: namely human-awareness~\cite{sisbot2007human,kruse2013human}.
	
	Inspired by the concept of egocentric and allocentric spatial cognition~\cite{proulx2016relation}, we propose a framework using ego-safety to measure collision from the robot's perspective and social-safety to measure the impact of our robot's actions on surrounding pedestrians. Specifically, social-safety is defined as the intrusion into surrounding human's interaction area considering a look-ahead distance (fig. 2A). We train a policy in our specially designed simulator with both randomized maps, obstacles, and moving agents of randomized numbers, sizes, shapes, and locations. Our learned policy is then directly evaluated in Gazebo and real robot test. Results show the learned policy demonstrates both time-efficient path planning behavior in a mapless navigation task and cooperative behavior which actively drives the robot into the crowd flows while showing respect to nearby pedestrians.
	
	\section{Related Works}
	This work is inspired by researches in the following topics.
	
	\textbf{Mapless Navigation}: The problem of mapless navigation~\cite{tai2017virtual,zhelo2018curiosity} studies how to autonomously navigate a robot in unknown environments given a predefined target. The assumption of accessibility to a 3rd party localization module is commonly applied, however limits its practical applications. For example, in outdoor scenarios, there are situations when GNSS signals are weak or denied by surroundings; in indoor scenarios, deploying a localization solution is tedious. Visual navigation~\cite{guerrero2005visual,zhu2017target,gao2017intention,chen2019behavioral} approaches remove the assumption, however, is challenging to learn dynamic collision avoidance behavior simply from visual inputs. One potential solution is a \textit{hierarchical approach}~\cite{gao2017intention} which has both a high-level global path planner and low-level motion controller. The high-level global path planner, is obtained from either a visual navigation module or simply GPS waypoints, that enables long-term navigation. The low-level motion controller learns both static and dynamic collision avoidance behavior to reach a local target defined by the global path planner. In this paper, we assume such high-level path planner for a long-term navigation is given, the focus is to learn a local navigation controller.
	
	
	\textbf{Social-aware/ Human-aware robot navigation}: Following the same approach as CADRL, \textit{social-aware} navigation problem is further considered which views more human-robot interactions in the navigation behavior. Chen et al.~\cite{chen2017SACADRL} propose SACADRL which considers human-like social norm behavior by adding a complex social norm reward. Similarly, Tai et al.~\cite{tai2018socially} train a social-compliant policy from RGB-D raw data inputs. These methods consider \textit{human-aware} robot navigation with social rules (norm) and human comfort. Apart from these complicated considerations, a simplified version is practical by only considering learning a cooperative ability to solve the freezing robot problem~\cite{fan2019getting}. Pfeiffer et al.~\cite{ETH2016MaxEnt} propose a framework to learn cooperative motion planning behavior by modeling human-robot interaction using maximum entropy methods. Chen et al.~\cite{chen2019crowdEPFL} jointly model human-robot and human-human interaction followed by a similar value function estimation in CADRL~\cite{chen2017CADRL}. Above methods depend on explicit pedestrian detection.
	
	\textbf{Navigation learning from observations}: Recently, learning the navigation strategy end-to-end directly from 2D laser scan readings proves possible. Long et al.~\cite{long2018towards} propose a framework to train a multi-agent collision avoidance policy using PPO~\cite{schulman2017proximal} which maps 3 frames of historical laser scans to continuous robot actions. They show the learned model can be further directly applied in dense crowd navigation tasks~\cite{fan2018crowdmove}. Apart from using 2D laser scans, navigation learning from visual inputs, a.k.a. visual navigation problem~\cite{zhu2017target,gao2017intention} has also been intensively studied.
	
	\begin{figure*}
	\centering
	\includegraphics[width=0.90\textwidth]{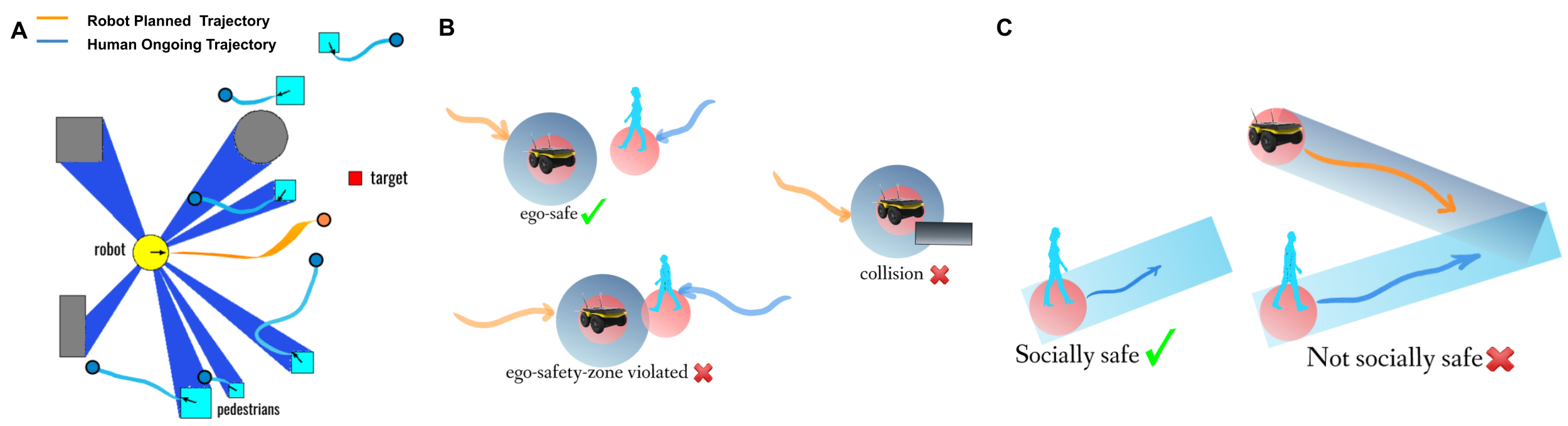}
	\caption{\textbf{A}: An example of robot and surrounding pedestrians' movements during a time window of 40 steps. \textbf{B}: Illustration of robot ego-safety zone. \textbf{C}: Illustration of pedestrian's social-safety zone.}	
	\label{fig:net_design}
\end{figure*}

\begin{figure}
		\begin{center}
			\includegraphics[width=0.42\textwidth]{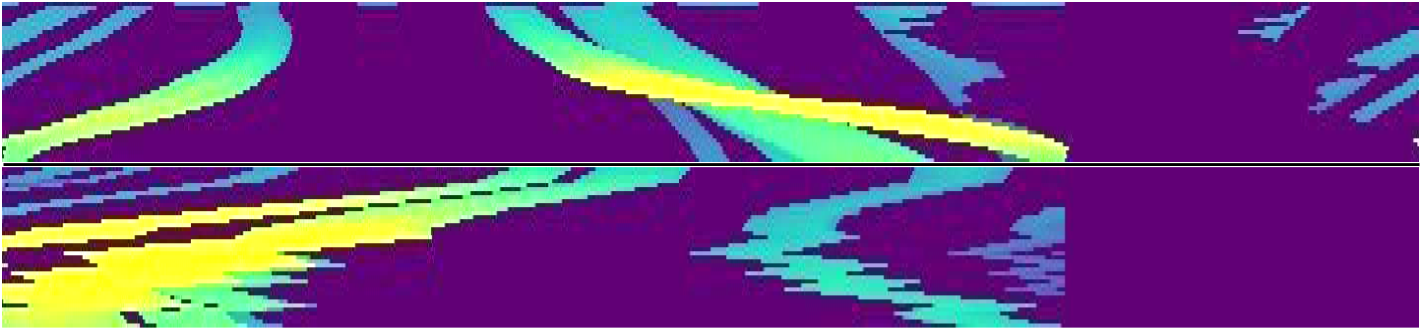} 
			\caption{Visualization of $\mathbf{o}^{t}_{m}$ with calibration (top) and without (bottom). When either the robot or a nearby pedestrian moves, the laser scan response will change, so will $\mathbf{o}^{t}_{m}$. We disentangle robot ego motion out to help the policy training.}
			\label{fig:design_overview}
		\end{center}
	\end{figure}
	
	\section{Approach}
	\subsection{Problem Formation}
	We formulate the problem as a Partially-Observable Markov Decision Process (POMDP) defined by a tuple $(\mathcal{S}, \mathcal{A}, \mathcal{P}, \mathcal{R}, \Omega, \mathcal{O})$, where $\mathcal{S}$ is the state space, $\mathcal{A}$ is the action space, $\mathcal{P}$ is the state transition model, $\mathcal{R}$ is the reward function, $\Omega$ is the observation space, $\mathcal{O}$ is the probability function defining how observations are obtained from the underlying environment state ($\mathbf{o}^t \sim \mathcal{O}(\mathbf{s}^t)$). The robot can only observe laser scan $\mathbf{o}^{t}_{z}$  from surroundings and relative location $\mathbf{o}^{t}_{g}$ to the target. 
	
	We define $\mathbf{o}^t=[\mathbf{o}^{t}_{m},\mathbf{o}^{t}_{g}]$. $\mathbf{o}^{t}_{m}$ is a motion feature which encodes the laser scan response changes along time axis due to surrounding motions. Robot action $\mathbf{a}^{t}$ is sampled from a stochastic policy $\mathbf{\pi}$ given observation $\mathbf{o}^t$: $\mathbf{a}^{t} \sim \mathbf{\pi}_{\theta}(\mathbf{a}^{t}|\mathbf{o}^{t})$, where $\mathbf{a}^{t}=[\mathbf{a}^{t}_{x}, \mathbf{a}^{t}_{y}]$ is within range [-1.5,1.5]. Considering a nonholonomic kinematics model, $\mathbf{a}^{t}$ is further converted to linear and angular velocity $[\mathbf{v}^{t}_{l},\mathbf{v}^{t}_{w}]$ with $\mathbf{v}^{t}_{l}=sqrt(\mathbf{a}^{t}_{x},\mathbf{a}^{t}_{y})$ and $\mathbf{v}^{t}_{w}=atan(\mathbf{a}^{t}_{y}/\mathbf{a}^{t}_{x}) \in [-\pi, \pi]$.
	
	\subsubsection{Motion features}
	$\mathbf{o}^{t}_{m}$ plays as a prediction feature to learn a cooperative path planning behavior among dynamics. We consider disentangling robot ego motion out while constructing $\mathbf{o}^{t}_{m}$ from historical laser scans $\{\mathbf{o}^{t-k}_{z},...,\mathbf{o}^{t}_{z}\}$. We observe laser scan response can be changed by either robot or surrounding motions. By disentangling, motion feature $\mathbf{o}^{t}_{m}$ will only encode surrounding motions, which helps our policy training. While there are advanced methods to detect dynamic objects from laser scans~\cite{mertz2013moving,wang2014laser}, we observe that during a small step duration, laser reading is affected mostly by rotation rather than translations. For the efficiency of simulator training, we simply calibrate previous laser scan $\mathbf{o}^{t-i}_{z}$ based on the difference between robot heading angles at time t-i and t by a shift operation in the scan array. Fig. 3 shows the comparison between calibrated and uncalibrated results.
	
	\subsubsection{Reward function}
	The reward at each step is obtained from system state $\textbf{S}_t$ which is fully accessible in simulator training. Our reward function is a sum of ego-safety $R_e$, social-safety $R_s$ and reaching target $R_g$ rewards:
	\begin{equation}
	    \mathcal{R}(\textbf{s}_t) = R_e(\textbf{s}_t) + R_s(\textbf{s}_t) + R_g(\textbf{s}_t)
	\end{equation}
	
	We define the ego-safety zone (fig. 2B) of the robot as a circle around with radius ($r_i + 0.4$), where $r_i$ is the physical dimension of the robot. Given a set of nearby pedestrians \{$p^{t}_{j}$\} and obstacles $\{B_{k}\}$, their closest distance to robot is $d^{t}$. $R_e(\textbf{s}_t)$ is defined as:
	\begin{equation}
	       R_e(\textbf{s}_t)=\begin{cases}
	           -10 & \text{if collision happens} \\
               -0.25*(1 - \frac{d^{t}}{r_i + 0.4}) & \text{else if } d^{t}<r_i + 0.4\\
               0 & \text{ otherwise}
            \end{cases}
	\end{equation}
	
	We define the social-safety zone (fig. 2C) of each pedestrian as their interaction region stretching along current moving direction with a minimum safe headway distance in $\Delta t$. For computational efficiency, only pedestrians within 5m are considered. This zone is represented as a rectangle defined in pedestrian's reference frame with $width=-r_{object}$ and $height=r_{object} / 2 + d_{min} + \Delta t.\textbf{v}_{object}$, where $r_{object}$ is the radius of pedestrian bounding circle, $d_{min} = 0.5m$ is the minimum safe distance, $\textbf{v}_{object}$ is the current speed. We set $\Delta t=0.77s$ in training. The safety zones are then projected back into the world frame. Then we also construct the ego robot's social-safety zone likewise, and check if it intersects with that of other moving pedestrians. If intersects, we count it as one violation. Now we define $R_s(\textbf{s}_t)$ as:
	\begin{equation}
        R_s(\textbf{s}_t) = -0.1 * \frac{\text{number of violations}}{\text{total number of pedestrians}}
    \end{equation}
	
	Let $p^{t}$ be position of robot and $p^{*}$ be the target. The last term $R_g(\textbf{s}_t)$ is defined as:
	\begin{equation}
	       R_g(\textbf{s}_t)=\begin{cases}
               +10 & \text{ if reach the target}\\
               -0.01* \frac{\|p^t - p^*\|}{\|p^0 - p^*\|} & \text{ otherwise}
            \end{cases}
	\end{equation}

	\subsection{Parameterization}
	Our special consideration in parameterization is to make our network less sensitive to object shapes, which are mostly encoded in data along column axis in $\mathbf{o}^{t}_{m}$. So we design a large kernel size on the column direction followed by a max-pooling layer. We use DDPG~\cite{lillicrap2015continuous} for training. Our actor and critic network structure are illustrated in fig. 4.
	
	\begin{figure}
		\begin{center}
			\includegraphics[width=0.42\textwidth]{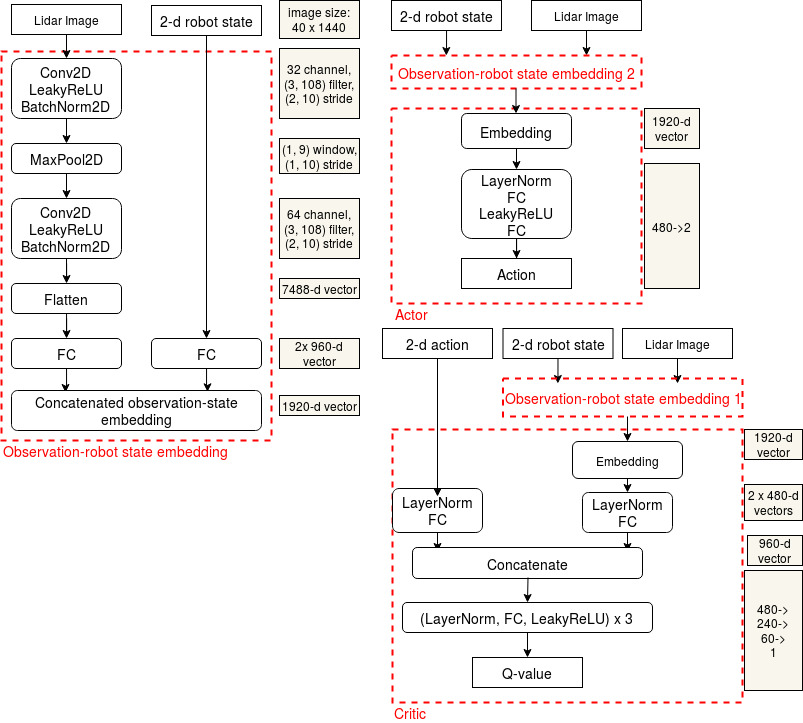} 
			\caption{Structure of our actor and critic network.}
			\label{fig:design_overview}
		\end{center}
	\end{figure}
	
	\subsection{Training the Policy}
	\subsubsection{Simulator}
	The policy is trained on our hand designed simulator (fig. 5) that runs a laser scanner at 40Hz and an inner differential drive controller at 20Hz. We model the laser scan simulation according to Hokuyo UTM-30LX with a $270\degree$ scanning range and 0.1 meter and 10 meters scanning distance. Historical 40 frames of laser scan are used to construct $\mathbf{o}^{t}_{m}$. The number, size and shape of static obstacles are randomized in simulation. Pedestrian behavior is also randomized with the following considerations: number of pedestrians in the scene, each pedestrian's current pose / velocity, each pedestrian's desired target and direction, each pedestrian's geometric shape and size, the behavior of stop-and-go, new walk-in pedestrians from random directions. Each pedestrian's behavior is controlled using ORCA~\cite{ORCA} but ignores the robot since otherwise pedestrians will always avoid the robot.
	
	\subsubsection{Training strategy}
	We train an \textit{ego policy} by only including the ego-safety reward and a \textit{social policy} by adding both ego-safety and social-safety rewards for comparison. The policy is trained using DDPG~\cite{lillicrap2015continuous}. To facilitate training, the social policy is trained based on learned ego policy parameters. 
    
	\section{Evaluation}
	
	\begin{figure}
	\setlength{\belowcaptionskip}{-10pt}
		\begin{center}
			\includegraphics[width=0.48\textwidth]{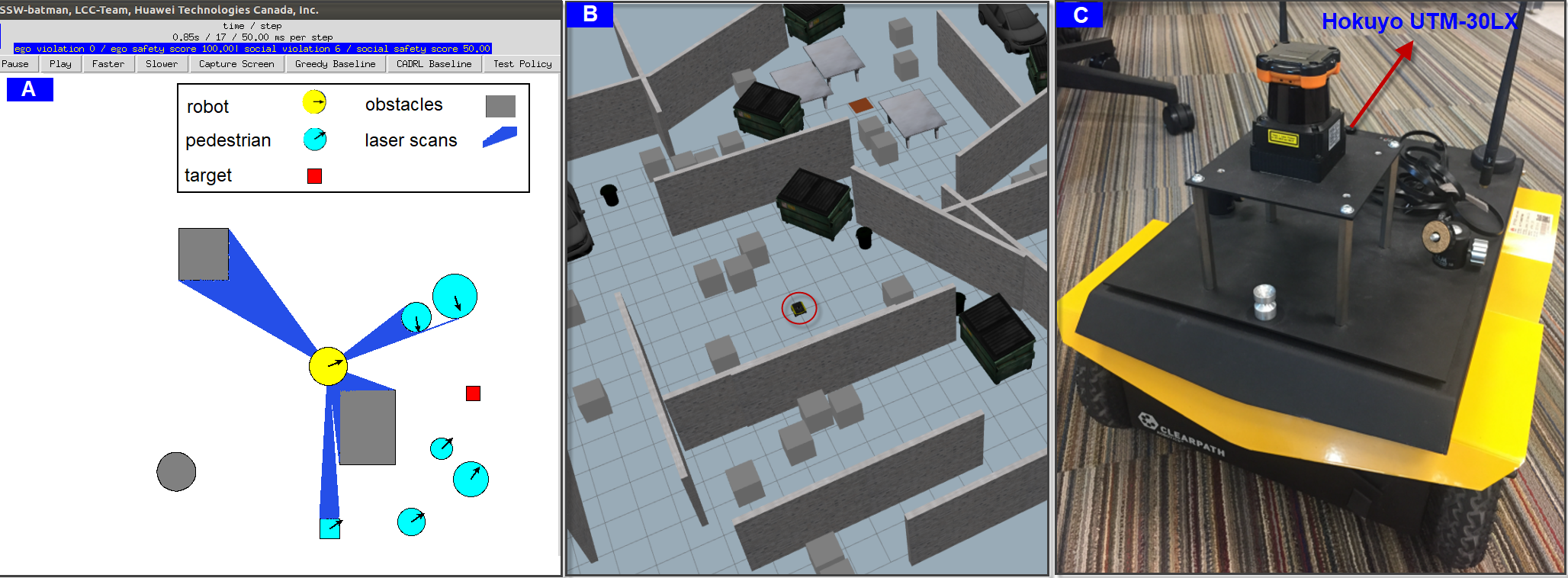} 
			\caption{The policy is trained in our specially designed simulator (\textbf{A}) and tested under various settings on our   simulator, Gazebo(\textbf{B}) and a Jackal robot(\textbf{C}) in real world environment respectively.}
			\label{fig:design_overview}
		\end{center}
	\end{figure}
	\begin{table}[]
\caption {Results of mapless path planning test in simulator. For each test, the robot needs to reach a fixed target in 10 randomly generated maps.} \label{tab:title} 
\begin{center}
\begin{tabular}{@{}cccc@{}}
\toprule
              & Success Rate & Arriving Time & Ego Score \\ \midrule
Greedy Baseline      & 80 \%        & 7.0 $\pm$ 1.6 s  & 100       \\
\textbf{(Ours)} Ego Policy    & 100 \%       & 3.1 $\pm$ 0.3 s  & 100       \\
\textbf{(Ours)} Social Policy & 100 \%       & 3.3 $\pm$ 1.3 s & 100       \\ \bottomrule
\end{tabular}
\end{center}
\end{table}
	\begin{figure}[h]
		\begin{center}
			\includegraphics[width=0.48\textwidth]{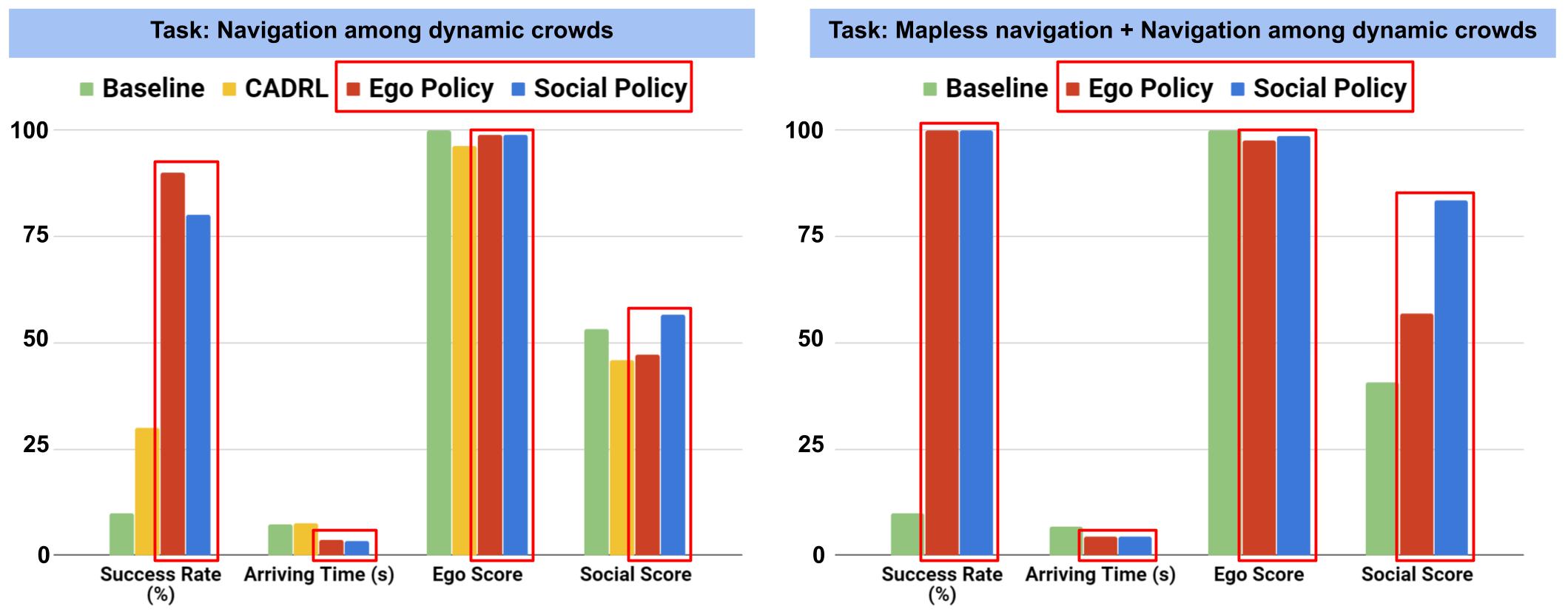} 
			\caption{Comparison between our method (ego policy, social policy) and the baseline in   simulator. \textbf{Left}: results on the navigation among dynamic crowds task. \textbf{right}: results of the combined task.}
			\label{fig:design_overview}
		\end{center}
	\end{figure}
     \begin{figure*}[]
	\setlength{\belowcaptionskip}{-10pt}
	\centering
	\includegraphics[width=0.95\textwidth]{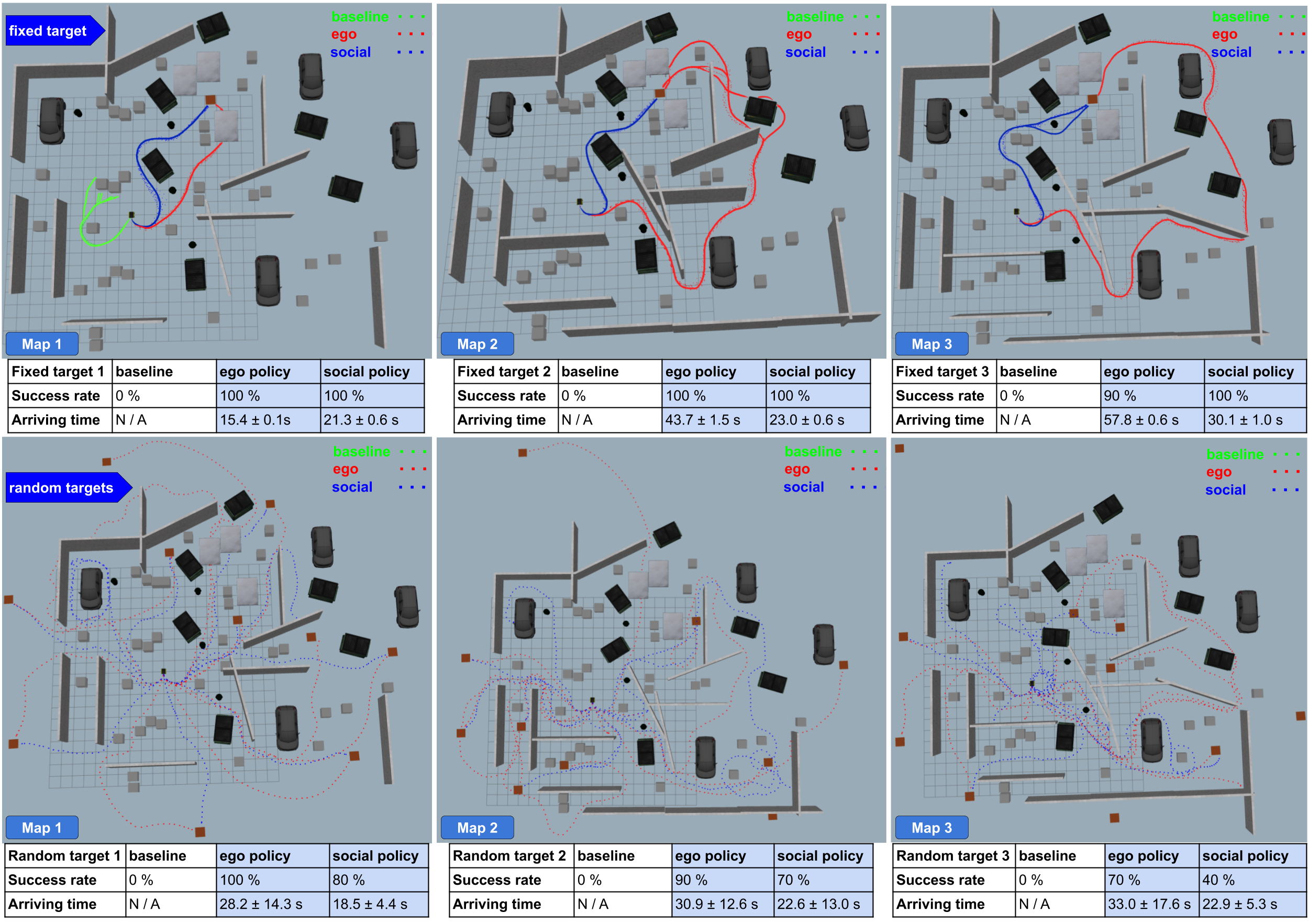}
	\caption{Evaluation results on the mapless path planning task in three Gazebo maps on a ROS environment. Map complexity increases from left to right. Up and down rows show result of a fixed target and a random target respectively.}
	\label{fig:net_design}
    \end{figure*}
    
	We evaluate the learned policy on \textit{three tasks} that all require the robot reaching a target: {(1)} Mapless path planning: The robot is required to find a path while avoiding obstacles in an unknown environment without pedestrians. {(2)} Navigation among dynamic crowds: Given a dynamic crowd with random behaviors, the robot needs to cross the crowd while showing awareness of social-safety without the `freezing robot' behavior. {(3)} Mapless path planning + navigation among dynamic crowds: A combination of the above two.
	
	\textbf{Baselines:} To the authors' best knowledge, there are no current methods that tackle both the mapless navigation and collision avoidance among dynamic crowds using 2D laser scans. However we still hand designed two baselines for comparison. {(1) The greedy baseline} takes input of laser scan and mimics human intuition on reaching the target while avoiding collisions locally. In the planner we take robot's direction and the difference of angle between the robot and the destination. We apply 1D convolution window with laser scan and find the index having the higher convoluted value with a p controller to turn the robot to the direction. {(2) The CADRL~\cite{chen2017CADRL} baseline} takes the full state of surrounding pedestrians as input. It's not designed for avoiding static obstacles in path planning, so it's only used in navigation among the dynamics crowds. Though it can't handle laser scan observations, we still feed in full states of all humans obtained from simulator, while testing our methods using only laser scans.

	\textbf{Metrics:} We designed {four metrics} for quantitative evaluation: {(1)} Success Rate (\%) after 10 runs in random settings. {(2)} Arriving Time (s). {(3)} Ego Score (0-100) to measure ego-safety awareness of robot. Let k be the number of ego-safety violation steps, and N be total steps to reach the target, $Ego\_Score=(1-k/N)*100$. {(4)} Social Score (0-100) to evaluate social-safety awareness of nearby human. Let m be the number of social-safety violation steps, $Social\_Score=(1-m/N)*100$.
	
	We compare {both} our ego policy and social policy to the above two baselines. Experiments are conducted in both our simulator and Gazebo for quantitative evaluation, while in real world tests for qualitative evaluation (fig. 3).
    \begin{figure*}
    \setlength{\belowcaptionskip}{-10pt}
	\centering
	\includegraphics[width=0.98\textwidth]{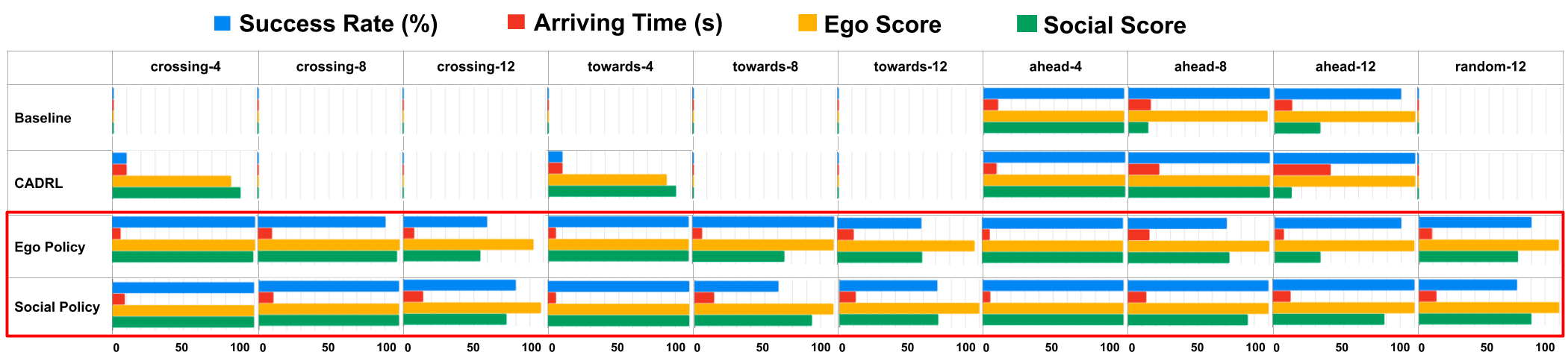}
	\caption{Comparison between our method and two baselines on navigation among dynamic crowds tasks in Gazebo.}
	\label{fig:net_design}
    \end{figure*}
	  \begin{figure}[h]
	  \setlength{\belowcaptionskip}{-10pt}
		\begin{center}
			\includegraphics[width=0.35\textwidth]{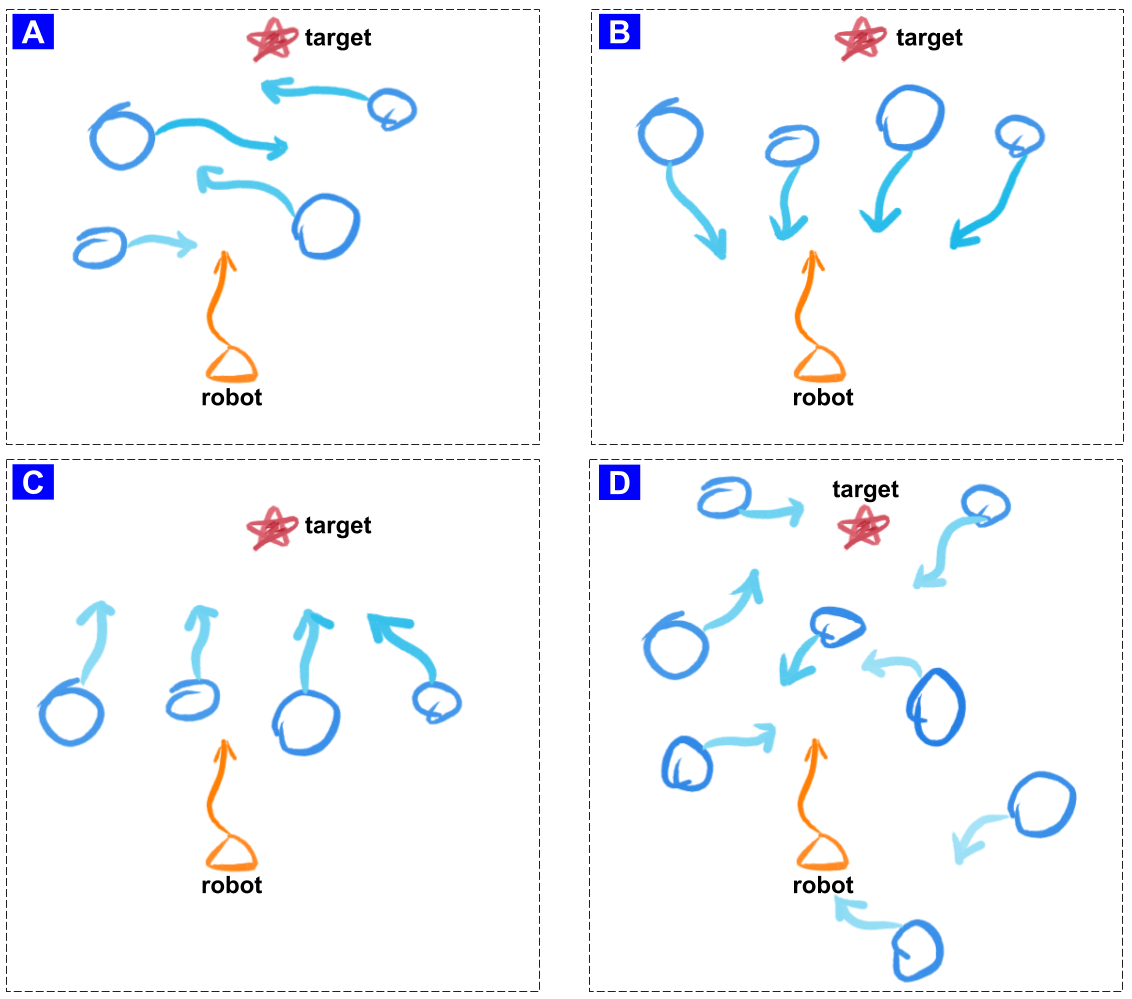} 
			\caption{The four crowd behavior scenarios. \textbf{A} crossing: The robot needs to move through the random crossing flow. \textbf{B} towards: Pedestrians are walking towards the robot. The robot needs to react fast since the relative velocity is large. \textbf{C} ahead: Pedestrians are walking ahead. The robot needs to decide how to pass. \textbf{D} random: All pedestrians' intention are randomized.}
			\label{fig:design_overview}
		\end{center}
	\end{figure}

    \begin{figure}
	\setlength{\belowcaptionskip}{-10pt}
		\begin{center}
			\includegraphics[width=0.4\textwidth]{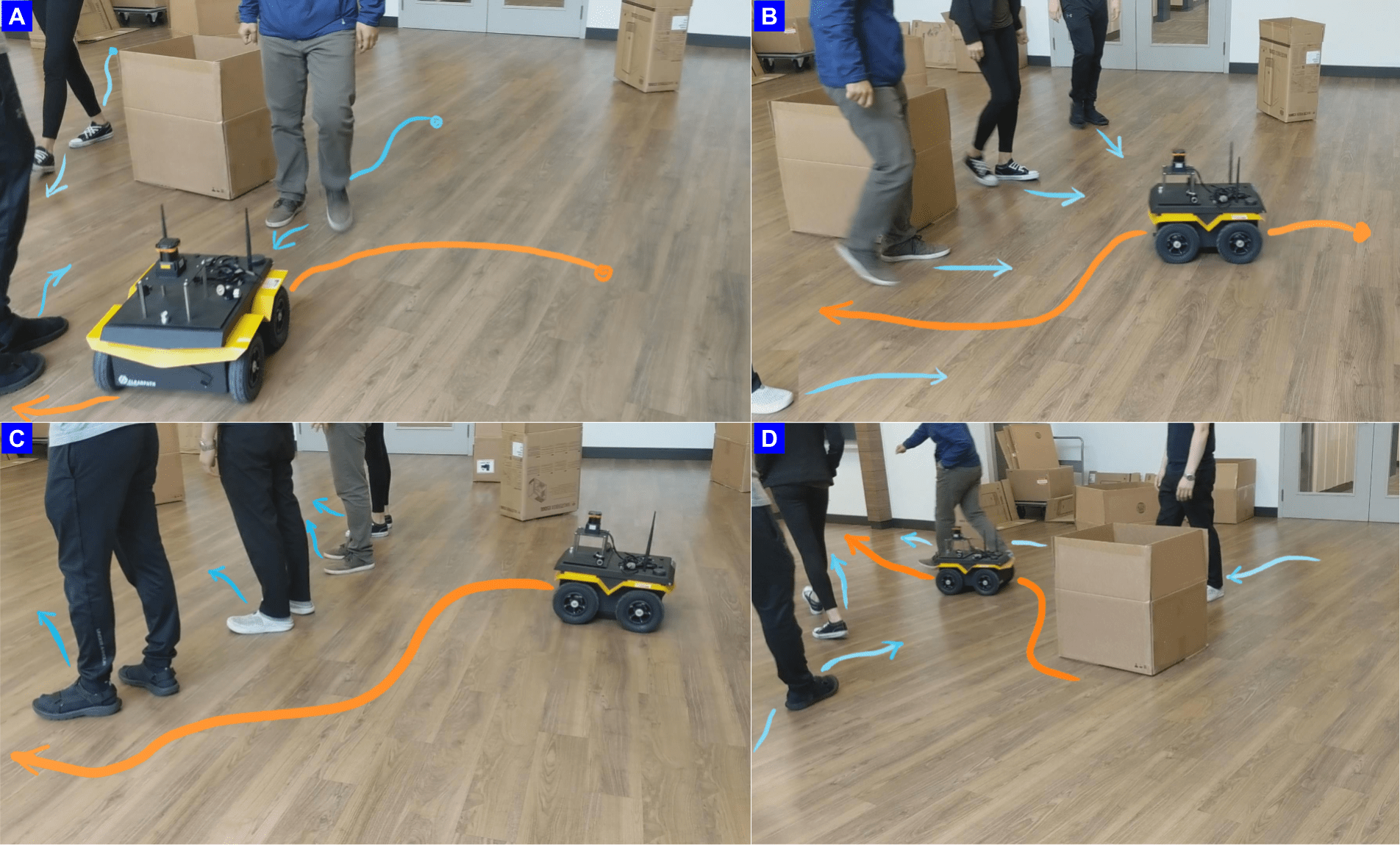} 
			\caption{Our observed robot behavior using social policy on the four crowd scenarios. \textbf{Blue} curve shows human trajectory; \textbf{Orange} curve shows robot trajectory. Our observation of robot behaviors in real world experiments aligns with our Gazebo test results. More evaluation results are included in our video.}
			\label{fig:design_overview}
		\end{center}
	\end{figure}
	
	\subsection{In the simulator}
	\subsubsection{Mapless path planning task}
	In this task, we fix the target while generating 10 random maps that only have static obstacles. Since CADRL can't handle this task, we only compare with greedy baseline (Table 1). Results show both our ego and social policy reach the target more efficiently with a higher success rate.
	
	\subsubsection{Navigation among dynamics task}
	We evaluate using 10 random crowds for each method. Each crowd has at least 8 pedestrians in a 5m$\times$5m area with a chance of walk-in humans from a random direction. All pedestrians have no sense of the robot so they won't avoid it. We compare our methods to both baselines (Fig. 6-left).
	
	\subsubsection{The combined task settings}
	Now we combine the above two task settings. Since CADRL can't work in mapless path planning, we only compare with the greedy baseline. Results are plotted in fig. 6-right. 
	
	Our results show the CADRL baseline typically has a passive `wait and go' behavior which is not efficient and easily get trapped by crowds. In comparison, both our policies show more cooperative behaviors. This cooperative behavior helps the robot reach target more efficiently with higher success rate. While all methods achieve high ego score, the social policy shows better performance in social-safety awareness. By observation, the ego policy is more aggressive compared to our social policy. Detailed videos are on our website.

	\subsection{In Gazebo environment}
	We used the Gazebo maps and the ROS platform for our experiments. 
	\subsubsection{Mapless path planning task}
	We evaluate the mapless navigation ability of our learned policy in three maps from simple to complex. Complexity increases from map 1 to map 3 by blocking key passages and adding more random obstacles. All tasks require the robot reaching the target (orange square) without knowing the map. We test in two scenarios: fixed target and random targets. Each scenario for each map runs 10 times on each method. Our results show the greedy baseline always fail as the map is too complex compared to simulator. Results (Fig. 7) show that all methods achieve ego score 100 in successful runs. We observe both ego and social policy show the behavior of slowing down when obstacles are highly aggregated. The ego policy performs better since it positively runs via small gaps between obstacles. However, it easily gets trapped into dead loops, where it will run circle loops instead of tracing back to find a way out. As comparison, the social policy is more cautious even towards static obstacles. It always runs into wider open spaces and won't take the risk to go through small gaps, in which case, it will switch to other directions. If the gaps in all directions are small, it will run a circle loop.
 
	\subsubsection{Navigation among dynamics crowds}
	We designed four random behaved crowd scenarios (Fig. 9) considering crowd size 4, 8, 12 in a 5m$\times$5m area. Each method on each scenario / crowd size combination needs to run 10 times to get evaluation results, as shown in Fig. 8. Results show our method reaches the target in a shorter time with a higher success rate. We observe the CADRL baseline prefers moving slowly to wait for crowds pass, however, may collide with human approaching from other directions. In comparison, our ego and social policies typically move into the crowd flow while avoiding pedestrians dynamically. In all tests, the social policy has the highest social score. Among the four scenarios, the `towards' one is most difficult. Our ego policy has the best performance since it's more agile running through small gaps. The `ahead' scenario is the easiest one. Both baselines simply move slowly behind the crowds. In contrast, our ego and social policy positively try to pass the crowd ahead. So they are significantly faster than the two baselines.

	\subsection{In real world tests}
We also conduct qualitative evaluations on a Jackal robot. It is worth noting the robot pose we used in training is accurate since it is obtained from the simulator. However, in our real world test, we don't have a precise 3rd party localization module due to resource limits. Using laser scan based SLAM localization\cite{KohlbrecherMeyerStrykKlingaufFlexibleSlamSystem2011} methods may help, but the localization output contains errors when running among dynamic human crowds. As a trade-off, we rely on robot's built-in odometry module which computes robot pose using wheel encoders and on-board IMU. To avoid drifting errors long time, we reset the odometry module after each run. We observe that, even with this coarse localization input, the robot performs relatively well. 

We tested two tasks in an indoor room. For each task, the robot needs to navigate to the target while avoiding obstacles and humans. For each run, we assign the target coordinates in the robot's initial odometry frame when we reset it. In general, the target is around 3 to 6 meters ahead of the robot.
	
	\subsubsection{Mapless path planning task}
	We design two maps for this task without pedestrians. In each map, the robot needs to navigate to the target while avoiding static obstacles.	We also test our method's performance while humans push obstacles into the robot's path. Evaluation results are shown in our video. For static obstacle avoidance in the two maps, both ego and social policies perform well. For pushing obstacles setting, the ego policy agilely avoids human kicking boxes while the social policy behaves more cautiously. 
	
	\subsubsection{Navigation among dynamics task}
	We also test our social policy in the four crowd scenarios (Fig. 9). Fig. 10 shows the observed robot behavior. Our empirical results show the policy generalizes from regular geometric shape based obstacles in simulation to human legs in real world. This is achieved by specially designed convolution filter and stride size on the historical laser scan observations $\mathbf{o}^{t}_{m}$. The row axis of $\mathbf{o}^{t}_{m}$ encodes changes along the time line, while the column axis encode changes because of geometry shapes of surrounding objects. We designed large convolution filter and stride size on the column axis while small scales on the row axis.
	
	In experiments we observed the circling behavior when the robot tries to pass an exit point out of an obstacle closure or avoid a near-close pedestrian, but restricted to its nonholonomic kinematics constraints. The policy learns to gradually change its twist velocities until precisely hit the exit point to escape the obstacle closure or avoid close pedestrians. Ideally, such behavior is not the most efficient one since the robot can simply stop, rotate and move. We leave it as a future problem to learn stop behaviors without sacrificing efficiency.

	\section{Conclusion}
	We propose a method to tackle the problem of mapless collision-avoidance navigation where humans are present using 2D laser scans by the formation of ego safety and social safety. Extensive experiments were conducted with both quantitative and qualitative evaluation. Results show our method can demonstrate cooperative path planning behavior by predicting the future motion of humans, while considering the impact of the robot actions on surrounding humans, namely the awareness of social-safety.
	
	Though some success is achieved in this project, other practical considerations are worth pointing out: (1) To define the target, mapless navigation relies on a third party localization module. This is usually unrealistic considering it is expected to work in unknown environments. As discussed in section 2, a high level abstract path planner with vision sensor support may solve the problem. (2) We find the training process is still tedious to fine tune the parameters. Though~\cite{long2018towards,fan2018crowdmove} propose a practical multi-stage training strategy which trains the policy from simple scenario to complex ones, hitherto, it lacks theoretical guarantees. We expect finite horizon reinforcement solutions may help since path planning problem essentially needs to consider multiple steps ahead.
	
	The authors would like to give thanks to all our colleagues participating in the real world tests.
	\addtolength{\textheight}{-2cm}   
	



	
	\bibliographystyle{IEEEtran}
	\bibliography{IEEEabrv,IEEEexample}
	
\end{document}